\pdfoutput=1

\documentclass[11pt]{article}

\usepackage{EMNLP2022}

\usepackage{times}
\usepackage{latexsym}
\usepackage{subfigure}
\usepackage[T1]{fontenc}

\usepackage[utf8]{inputenc}

\usepackage{microtype}

\usepackage{inconsolata}

\usepackage{soul}
\usepackage{multirow}
\usepackage{amssymb}
\usepackage{pifont}
\newcommand{\cmark}{\ding{51}}%
\newcommand{\xmark}{\ding{55}}%
\usepackage{graphicx}
\newcommand\Tstrut{\rule{0pt}{2.6ex}}         
\newcommand\Bstrut{\rule[-0.9ex]{0pt}{0pt}}   
\usepackage{mathtools,amsthm}
\DeclareMathOperator*{\argmax}{arg\,max}
\DeclareMathOperator*{\argmin}{arg\,min}
\usepackage{bm}
\usepackage[ruled,vlined,linesnumbered]{algorithm2e}
\SetKwInput{kwInit}{Initialize}

\newcommand{\shortname}{OPTIMA}
\newcommand{\fullname}{bOosting Prompt TunIng with doMain Adaptation (OPTIMA)}
\newcommand{\expect}{\mathop{\mathbb{E}}}
\newcommand{\iid}{\overset{\mathrm{i.i.d.}}{\sim}}
\usepackage{mathtools}

\DeclarePairedDelimiterX{\infdivx}[2]{(}{)}{%
  #1\;\delimsize\|\;#2%
}
\newcommand{\KLdiv}{\mathrm{KL}\infdivx}
\DeclarePairedDelimiter{\norm}{\lVert}{\rVert}

\title{Improving the Sample Efficiency of Prompt Tuning with Domain Adaptation 
}

\author{Xu Guo, Boyang Li, \and Han Yu\\
  School of Computer Science and Engineering,\\
  Nanyang Technological University, Singapore\\
  \texttt{\{xu008, boyang.li, han.yu\}@ntu.edu.sg} \\
}

\begin{document}
\maketitle
\begin{abstract}
Prompt tuning, or the conditioning of a frozen pretrained language model (PLM) with soft prompts learned from data, has demonstrated impressive performance on a wide range of NLP tasks. However, prompt tuning requires a large training dataset to be effective and is outperformed by finetuning the entire PLM in data-scarce regimes. 
Previous work \cite{gu-etal-2022-ppt,vu-etal-2022-spot} proposed to transfer soft prompts pretrained on the source domain to the target domain. 
In this paper, we explore domain adaptation for prompt tuning, a problem setting where unlabeled data from the target domain are available during pretraining.
We propose \fullname{}, which regularizes the decision boundary to be smooth around regions where source and target data distributions are similar. Extensive experiments demonstrate that \shortname{} significantly enhances the transferability and sample-efficiency of prompt tuning compared to strong baselines. Moreover, in few-shot settings, \shortname{} exceeds full-model tuning by a large margin.

\end{abstract}

\section{Introduction}

Prompt tuning \citep{lester-etal-2021-power,li-liang-2021-prefix,liu-etal-2022-p,hambardzumyan-etal-2021-warp} is an effective method for adapting large-scale pretrained language models for downstream tasks. 
While keeping the PLM weights unchanged, prompt tuning trains input vectors, called soft prompts, that are input to the PLM alongside the text embeddings. 
Compared to other adaptation techniques for PLMs, such as Adapter \cite{houlsby2019adapter,ruckle-etal-2021-adapterdrop,he2022parallel-adapter}, Compacter \cite{karimi2021compacter}, BitFit \cite{ben-zaken-etal-2022-bitfit}, LoRA \cite{hu2021lora}, and Ladder Side-Tuning \cite{sung2022-ladder}, the advantage of prompt tuning is that it does not require addition or change of model parameters. As a result, with prompt tuning, we can easily specialize one neural network (possibly deployed on a large number of servers or as application-specific integrated circuits) to support many different tasks by simply switching out the soft prompt in the input, which greatly simplifies model deployment and maintenance.

\begin{figure}
\centering
\includegraphics[width=\columnwidth]{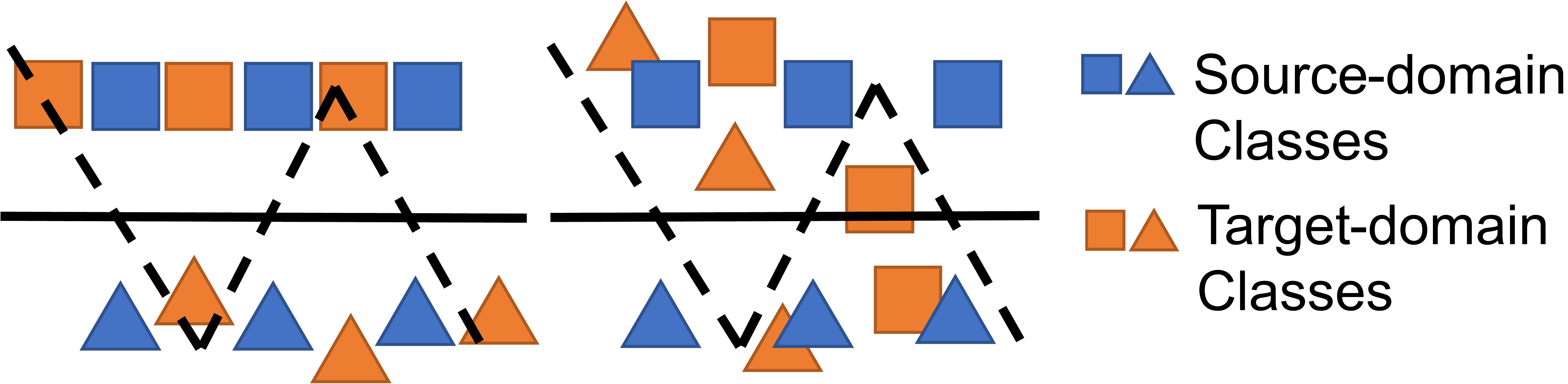}
\caption{Smooth vs. zigzag decision boundaries. Left: When the distribution of the target-domain data (orange) are similar to the source domain (blue), the smooth decision boundary (solid line) generalizes better than the zigzag boundary. Right: When the distributions are different, it is not clear if the smooth decision boundary is the better choice. }
\label{fig:intuition}
\end{figure}

However, training effective soft prompts usually requires sufficient labeled training data \cite{su2021transferability}. Studies have shown that prompt tuning significantly underperforms full-model tuning on many few-shot classification tasks \cite{gu-etal-2022-ppt}. Our experiments corroborate this finding. In addition, we find that, in few-shot learning, prompt tuning is equally, if not more, sensitive to random seed choices compared to full-model tuning, despite having far fewer trainable parameters (\S \ref{exp:few-shot-with-init}). 
\citet{gu-etal-2022-ppt} address this by transferring prompts learned from a source domain to the target domain with limited training data. 


In this paper, we investigate a related but different scenario, unsupervised domain adaptation (UDA) \cite{wang-etal-2019-adversarial,long-etal-2022-domain}, where unlabeled data from the target domain are available. Such situations are common when data are abundant but the labeling cost, including annotator recruitment, annotator training, and quality assurance, is high. Utilizing  unlabeled examples can be an effective approach toward enhancing the data efficiency of prompt tuning. 


We propose \fullname{}. Employing regularization from adversarial perturbation, \shortname{} learns a smooth decision boundary that passes through regions of low data density. In addition, recognizing that the feature distributions in the two domains may overlap only partially, we propose to focus the regularization to regions where the target-domain and source-domain data exhibit high similarity. We illustrate the intuition in Figure \ref{fig:intuition}. 

The popular Domain-adversarial Neural Network (DANN) technique \cite{dann-2016}  encourages the network to learn domain-invariant features and optimizes for both a domain-specific task loss and a domain discrimination loss. However, the two losses could compete against each other, leading to optimization difficulties \cite{guo-etal-2021-latent}. Empirically, DANN exhibit low performance for prompt tuning. We hypothesize that the low capacity of prompts worsens the optimization problem. To solve this issue, in \shortname{}, we create input disturbance vectors 
that optimize for domain similarity, so that the prompt needs to optimize for only the task loss. The separation leads to excellent results.

Experiments shows that \shortname{} learns effective data representations that transfer well to the target domain under zero-shot and few-shot settings. \shortname{} outperforms eight baselines, including state-of-the-art transfer learning techniques such as SPOT \citep{vu-etal-2022-spot}. 
%
Our contributions include the following. 
\begin{enumerate}
    \item To our best knowledge, \shortname{} is the first domain adaptation technique for soft prompt tuning, which does not require any labeled data from the target domain. Empirical results show that the unlabeled data boost target-domain performance significantly. 
    \item Catering to partial overlaps of the data distributions, we propose a targeted regularization technique that encourages smooth decision boundaries only in the areas where the two domains are similar. 
    \item Through empirical evaluation, we show that \shortname{} outperforms state-of-the-art baselines,  improves data efficiency significantly, and effectively addresses domain shifts. Code and data are available at \url{https://github.com/guoxuxu/soft-prompt-transfer/tree/main/optima}.
\end{enumerate}

\section{Domain Adaptation for Prompt Tuning}

In this section, we first introduce prompt tuning for text classification. Then, we introduce how to enhance in-domain generalization performance of soft prompts by augmenting the input with virtual perturbations. Next, we propose how to optimize the perturbations to reduce the domain gap and obtain soft prompts with domain-invariant knowledge. Finally, we show how to use the soft prompts to boost few-shot learning in the target domain. 

\subsection{Preliminaries: Prompt Tuning} 
\label{sec:prompt-tuning}
We start by introducing some notations. 
The input $\bm x$ is a sequence of $n$ token embeddings, $\bm x = \langle x_1, \ldots, x_n \rangle$. The trainable soft prompt sequence $\bm p$ has $m$ embeddings, $\bm p = \langle p_1, \ldots, p_m \rangle$. The manually designed hard prompt sequence $\bm h$ has $k$ token embeddings $\bm h = \langle h_1, \ldots, h_k \rangle$. All embedding vectors have $d$ dimensions. The soft prompt and the hard prompt are both task-specific. The hard prompt text is usually a natural language description of the task, whereas the soft prompts do not correspond to any text and are trained directly using gradient descent. 

For classification problems, we adopt the masked language modeling formulation, which aims to predict a predefined verbalizer token $y \in \mathcal{V}$ at a masked position in the input. For example, for binary classification, the words ``yes'' and ``no'' may be used as verbalizers that indicate positive and negative predictions, where we may define the label space as $\mathcal{Y}=\{\mathrm{yes}, \mathrm{no}\}$. In encoder-only networks such as BERT \cite{devlin-etal-2019-bert}, the output of the encoder is mapped to the label space $\mathcal{Y}$ via a projection head. In encoder-decoder networks like T5 \citep{2020t5}, the decoder is responsible for generating the verbalizer token. 

We concatenate all sequences and the embedding of the [MASK] token, $e([\textrm{MASK}])$, to form the final input to the PLM: $\langle \bm p; \bm h; \bm x; e[\textrm{MASK}]\rangle$.
For simplicity, we use the function $f(\bm x, \bm p)$ to denote the PLM prediction at the masked position, which is a multinomial distribution over $\mathcal{Y}$. 
We adopt the the cross-entropy classification loss $\ell_{\text{xe}}$ with the ground-truth label $y \in \mathcal{Y}$.
\begin{align}
\ell_{\text{xe}}(\bm x, y, \bm p) =- \mathrm{log}\ P(f(\bm x, \bm p)=y).
\end{align}
We optimize the soft prompt by minimizing the expected loss over the labeled training set, $\mathcal{D}$:
\begin{align}
\bm p^* = \argmin_{\bm p} \expect_{(\bm x, y)\in\mathcal{D}} \big[\ell_{\text{xe}}(\bm x, y, \bm p)\big].
\end{align}

\subsection{The \shortname{} Approach}

We build \shortname{} off two intuitions regarding domain adaptation. First, as the target domain provides no direct supervision, it is easy to overfit to the source domain. Therefore, it is important to mitigate overfitting by regularizing the network to maintain a smooth decision boundary. 

Under an adversarial learning framework, we seek a small perturbation $\bm \delta$ that, when added to the input, results in maximum change in the model prediction. After that, we optimize the model parameters to minimize the prediction change under the adversarially perturbed input. The overall result is a network whose output $f(\bm x)$ changes little where a small change is added to the input $\bm x$. In the sense of Lipschitz continuity, such a decision boundary is smooth. Smooth decision boundaries can be understood as passing through regions of low data density and are shown to improve generalization \citep{huang2020understanding,Cicek_2019_ICCV,kim2019understanding}. 

The second intuition is that we do not have to regularize the entire decision boundary. As the source and target domains may have different data distributions, all that matters is the decision boundary segment close to the target-domain data. Therefore, we target the regularization and the perturbation $\bm \delta$ to areas on the data manifold where the source domain and target domain are similar. 





\begin{figure}[t]
\centering
\includegraphics[width=0.6\columnwidth]{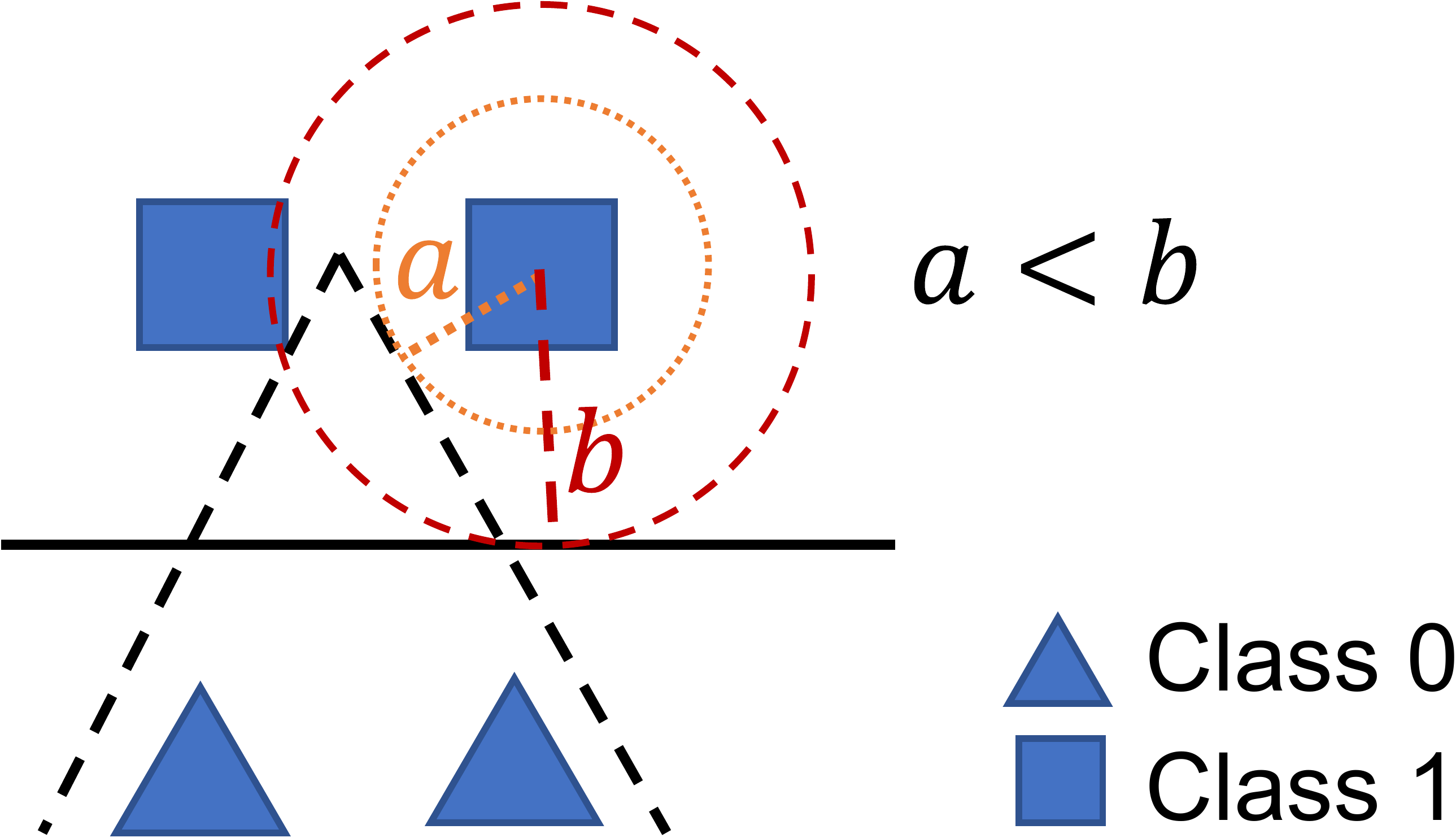}
\caption{Intuition about perturbation and smoothness. Under the zigzag (non-smooth) decision boundary, a small perturbation with a well-chosen direction is sufficient to flip the predicted class. The smooth boundary requires a larger perturbation. }
\label{fig:intuition2}
\end{figure}

Specifically, we have a labeled dataset from the source domain,  $\mathcal{D}_s=\{(\bm x_s^{(i)}, y_s^{(i)})\}_{i=1}^{N_s}$, drawn i.i.d.\ from distribution $P_{s}$ and an unlabeled dataset from the target domain, $\mathcal{D}_t=\{\bm x_t^{(j)}\}_{j=1}^{N_t}$, drawn i.i.d.\ from distribution $P_{t}$. We define $\ell_{\text{KL}}$ as the KL divergence between the prediction of the original input and that of the perturbed input, 
\begin{equation}
\label{eq:loss_rough}
\ell_{\text{KL}}(\bm \delta, \bm p, \bm x_s) = \KLdiv{f(\bm x_s, \bm p)}{f(\bm x_s + \bm \delta, \bm p)}.
\end{equation}
$\ell_{\text{KL}}$ measures how much the model prediction changes when the perturbation $\bm \delta$ is applied to $\bm x_s$ and captures the smoothness of the decision boundary.  We illustrate the intuition in Figure \ref{fig:intuition2}. 

Further, we introduce a domain discriminator network parameterized by $\bm \theta_{d}$, which attempts to distinguish data instances from the two domains. This network is trained to reduce the domain discrimination loss $\mathcal{L}_{\text{disc}}$,
\begin{align}
\mathcal{L}_{\text{disc}}(\bm \delta, \bm x_s, \bm x_t) = & \expect_{\bm{x}_s, \bm x_t} \big[  -\mathrm{log} \ P(z=1|\bm x_s + \bm\delta) \nonumber \\
& -\mathrm{log} \ P(z=1|\bm x_s) \label{eq:L_d} \\
& -\mathrm{log} \ P(z=0|\bm x_t) \big], \nonumber
\end{align}
where $z$ is the output of the discriminator network. This loss is a variation of the cross entropy with an additional term where $\bm x_s$ is perturbed by $\bm \delta$. 
In addition, we define an adversarial loss,
\begin{equation}\label{eq:L_adv}
\ell_{\text{adv}}(\bm \delta, \bm x_s) = - \mathrm{log} \ P(z=1|\bm x_s + \bm\delta), 
\end{equation}
which, when maximized, causes the domain discriminator to mistake the perturbed source example $\bm x_s + \bm\delta$ as coming from the target domain. 

For a given source-domain input, $\bm x_s$, we find the perturbation $\bm \delta^*$ within a $\epsilon$-radius ball that maximizes the following objective,
\begin{equation}
\bm \delta^* = \argmax_{\bm\delta,\|\bm\delta\|_2\leq\epsilon} \ell_{\text{KL}}(\bm \delta, \bm p, \bm x_s) + \ell_{\text{adv}}(\bm \delta, \bm x_s).
\label{eq:optimal-perturbation}
\end{equation}
Here, $\ell_{\text{adv}}(\bm \delta, \bm x_s)$ can be understood as a regularization term for $\bm \delta$. By maximizing $\ell_{\text{KL}}$, we seek a disturbance to the input that causes the most change in the model prediction. At the same time, the disturbed input $\bm x_s + \bm\delta^*$ from the source domain should resemble data in the target domain, in order to maximize $\ell_{\text{adv}}(\bm \delta, \bm x_s)$; $\ell_{\text{adv}}$ constrains $\bm \delta^*$ to the region where the data from the two domains are similar.

We optimize the above loss w.r.t.\ $\bm \delta$ using projected gradient ascent (PGA). After every gradient descent step, $\bm \delta$ is projected back to the $\epsilon$-radius ball $\mathcal{Q}_{\epsilon} = \{\bm \delta|\Vert\bm\delta\Vert_{2}\leq\epsilon \}$. We write the projection operation as
\begin{equation}
\prod_{\Vert \cdot \Vert_{2}\leq\epsilon}(\bm{\phi}) = \argmin_{\bm \delta \in \mathcal{Q}_{\epsilon} } \|\bm \delta - \bm \phi \|_2 = \frac{\epsilon \bm \phi}{\max(\epsilon, \|\bm \phi\|_2)}.
\end{equation} 
The update to $\bm \delta$ can be written as
\begin{align}
\bm \delta  & \leftarrow \prod_{\Vert \cdot \Vert_{2}\leq\epsilon}\left(\bm \delta + \eta_{\delta} \frac{\bm g}{\norm{\bm g}_2} \right), \\
\bm g & = \bigtriangledown_{\bm \delta}(\ell_{\text{KL}}(\bm \delta, \bm p, \bm x_s) + \ell_{\text{adv}}(\bm \delta, \bm x_s)),
\end{align} 
where $\eta_{\delta}$ is the learning rate. We normalize $\bm g$ to make sure the updates have the same magnitude. 


During the training session, we alternately optimize the perturbation $\bm \delta$ and the soft prompt $\bm p$. With $\bm{\delta}^*$ found by PGA, we optimize the following loss function over $\bm p$ using standard gradient-based optimization. 
\begin{align}
\mathcal{L}_{R} & = \expect_{(\bm x_s, y_s)\in\mathcal{D}_s} \big[\ell_{\text{xe}}(\bm x_s, y_s, \bm p) + \ell_{\text{KL}}(\bm{\delta}^*, \bm p, \bm x_s)\big] \nonumber \\
\bm p^* & = \argmin_{\bm p}\mathcal{L}_{R} \label{eq:tvat} 
\end{align}
$\mathcal{L}_{R}$ is the empirical expectation computed over the current mini-batch. With the same $\bm{\delta}^*$, we also minimize the domain discrimination loss over the discriminator network parameter $\bm \theta_{d}$. 

\subsection{The \shortname{} Algorithm}
We show the complete \shortname{} algorithm as Algorithm \ref{algo}. With lines 5 and 6, we create an initial perturbation $\bm \delta_0^{(i)}$ for every source data point $\bm x_s^{(i)}$. From line 7 to line 13, we iteratively update the perturbation $\bm \delta^{(i)}$ associated with every source-domain data point $\bm x_s^{(i)}$ using projected gradient ascent on $\ell_{\text{KL}} + \ell_{\text{adv}}$. After $K$ iterations, we find $\bm{\delta}^{(i)}* = \bm{\delta}^{(i)}_{K}$, compute $\bigtriangledown_{\bm p}\mathcal{L}_{R}$ accordingly, and update $\bm p$ with stochastic gradient descent (SGD) and learning rate $\eta_p$ (line 16).  At line 17, we update the domain discriminator parameters $\bm \theta_d$ using SGD with the current mini-batches. Though we show the vanilla SGD updates in lines 16-17, we can easily switch to other optimizers such as SGD with momentum or Adam \citep{DBLP:journals/corr/KingmaB14}.

\begin{algorithm}[t]\small
\SetAlgoLined
\KwIn{A labeled source-domain dataset $\mathcal{D}_s=\{(\bm x_s^{(i)} y_s^{(i)})\}_{i=1}^{N_s}$ and an unlabeled target-domain dataset $\mathcal{D}_t=\{\bm x_t^{(j)}\}_{j=1}^{N_t}$, perturbation ball radius $\epsilon$, ascent steps $K$ and step size $\eta_{\delta}$.}
\kwInit{Soft prompts embeddings $\bm p$ and domain discriminator $\theta_d$,  learning rates $\eta_p, \eta_d$.}

\Repeat{$\textrm{\upshape the maximum training epoch is reached}$}{
    
Sample a pair of batches, each of $B$ data points, from $\mathcal{D}_s$ and $\mathcal{D}_t$\;
\For{$i=0,...,B$}{
Forward computation: $f(\bm x_s^{(i)}, \bm p), \; \forall \bm x_s^{(i)}$ \\
Sample a $\bm\delta_{0}^{(i)} \sim \mathcal{U}(-1,1)$, $\forall \bm x_s^{(i)}$\\
$\bm\delta_{0}^{(i)} \leftarrow \prod_{\Vert \cdot \Vert_{2}\leq\epsilon}(\bm\delta_{0}^{(i)})$ \\
\For{$t=0,...,K-1$}{
    Forward with $\bm\delta_t^{(i)}$: $f(\bm x_s^{(i)} + \bm \delta_t, \bm p)$ \\
    Compute $\ell_{\text{KL}}(\bm \delta_t^{(i)}, \bm p)$ (Eq. \ref{eq:loss_rough})\\
    Compute $\ell_{\text{adv}}(\bm \delta_t^{(i)})$ (Eq. \ref{eq:L_adv})\\
    Perform PGA on $\bm\delta_t^{(i)}$: \\
    $\,\,\,\, \bm g \leftarrow \bigtriangledown_{\bm\delta_{t}^{(i)}}(\ell_{\text{KL}}(\bm \delta_t^{(i)}, \bm p) + \ell_{\text{disc}}(\bm \delta_t^{(i)}))$ \\
    $\,\,\,\, \bm\delta_{t+1}^{(i)} \leftarrow \prod_{\Vert \cdot \Vert_{2}\leq\epsilon} (\bm\delta_t^{(i)} + \eta_{\delta} \cdot \frac{\bm g}{\Vert \bm g\Vert_{2}})$\\
}
}
Compute $\mathcal{L}_{R}$ (Eq. \ref{eq:tvat}), $\mathcal{L}_{\text{disc}}$ (Eq. \ref{eq:L_d}) with $\bm\delta_{K}$ \\
$\bm p \leftarrow \bm p - \eta_p \bigtriangledown_{\bm p}\mathcal{L}_{R}(\bm x_s, y_s, \bm p)$ \\
$\bm \theta_d \leftarrow \bm \theta_d - \eta_d\bigtriangledown_{\bm  \theta_d}\mathcal{L}_{\text{disc}}(\bm\delta_{K}, \bm x_s, \bm x_t; \bm  \theta_d)$ \\
}
\KwOut{Learned soft prompt $\bm p$}
\caption{\shortname}
\label{algo}
\end{algorithm}

\subsection{Comparison with Virtual Adversarial Training}
Virtual Adversarial Training (VAT) \cite{miyato2016adversarial,miyato2018virtual} is a pioneering work that applies adversarial perturbation to unlabeled examples in semi-supervised learning (SSL). The SSL assumption is that we have labeled data $(\bm x, y) \iid P$ and unlabeled data $\bm x \iid P$. Notice that $\bm x$ is drawn from the same distribution $P$ regardless of the existence of the label $y$. VAT finds disturbance $\bm \delta \in \mathcal{Q}_{\epsilon}$ that maximizes the change in the model prediction $\KLdiv{f(\bm x)}{f(\bm x + \bm \delta)}$. After that, the neural network minimizes cross-entropy on labeled data and the KL-divergence under disturbance on all data. Similar ideas have been explored in \cite{Cicek_2019_ICCV, kim2019understanding,park2021consistency}.

A critical difference between SSL and domain adaptation is that the unlabeled data are drawn from a different distribution ($P_t$) than the labeled data ($P_s$). As the two distributions may overlap in some regions and diverge in others, regularizing over the the entire source dataset may be ineffective.
Thus, we propose to focus the smoothness constraint on the regions of the data manifold where the source-domain and target-domain data are similar. 

\section{Experimental Evaluation}
We evaluate the representations learned by \shortname{} under zero-shot and  few-shot settings. 

\begin{table}[t]
\centering
\resizebox{\columnwidth}{!}{%
\begin{tabular}{@{}ccccc@{}}
\hline \Tstrut
Dataset & Train & Test & $n_{class}$ & Verbalizers \Bstrut\\
\hline \Tstrut
MRPC & 4,076 & 408 & 2 & Yes/No \\
QQP & 363,847 & 40,430 & 2 & Yes/No \Bstrut\\
\hline \Tstrut
MNLI & 392,702 & 9,815 & 3 & Yes/Neutral/No \\
SNLI & 549,367 & 9,842 & 3 & Yes/Neutral/No\\
SICK & 4,439 & 4,906 & 3 & Yes/Neutral/No \\
CB & 250 & 56 & 3 & Yes/Neutral/No \Bstrut\\
\hline
\end{tabular}
}
\caption{Dataset characteristics. }
\label{tab:dataset}
\end{table}

\begin{table}[t]
\centering
\resizebox{\columnwidth}{!}{%
\begin{tabular}{@{}ccc@{}}
\hline \Tstrut
Paraphrase & NLI from MNLI & NLI from SNLI \Bstrut\\
\hline \Tstrut
MRPC $\rightarrow$ QQP & MNLI $\rightarrow$ SNLI & SNLI $\rightarrow$ MNLI   \\
QQP$\rightarrow$ MRPC & MNLI $\rightarrow$ SICK & SNLI $\rightarrow$ SICK \\
 & MNLI $\rightarrow$ CB & SNLI $\rightarrow$ CB \Bstrut\\
\hline
\end{tabular}
}
\caption{The set of domain adaptation experiments. }
\label{tab:test-suite}
\end{table}

\subsection{Datasets}
We investigate domain adaptation on six text classification datasets in two tasks. In the task of paraphrase detection, we employ MRPC and QQP\footnote{\url{https://quoradata.quora.com/First-Quora-Dataset-Release-Question-Pairs}}. In the task of natural language inference, we employ four datasets, including MNLI \citep{N18-1101}, SNLI \citep{bowman-etal-2015-large}, CB \citep{cb2019} and SICK \citep{marelli-etal-2014-sick}. The statistics and the label space $\mathcal{Y}$ of each dataset can be found in Table \ref{tab:dataset}. 
We prepare 8 groups of cross-domain experiments, two for  paraphrase detection and 6 for natural language inference (NLI), 
as shown in Table \ref{tab:test-suite}.

\subsection{Baseline Techniques}
We include eight competitive single-domain and cross-domain baselines. Out of the eight, baselines \#2-\#4 do not use any transfer learning from the source domain. Baselines \#5-\#9 utilize transfer learning and data from the source domain. 

\vspace{0.05in}
\noindent\textbf{1) Frozen PLM.}
Large PLMs have demonstrated non-trivial zero-shot performance \cite{brown2020language}. 
Here, we directly apply T5-large \cite{2020t5} with the manually written hard prompt and take the verbalizer with the highest probability as the prediction.

\vspace{0.05in}
\noindent\textbf{2) Prompt Tuning (PT).}
We feed the input data with both soft and hard prompts to a frozen T5-large model and finetune the soft prompt embeddings on the few-shot training set from the target domain.

\vspace{0.05in}
\noindent\textbf{3) Fine Tuning (FT).} We feed the input data with the hard prompt to T5-large and finetune the entire network on the few-shot target-domain data. Notice that we use the verbalizer rather than training a separate task-specific prediction head.


\vspace{0.05in}
\noindent\textbf{4) Prompt-based Fine Tuning (PFT).}
A representative method on exploiting soft prompts for fine-tuning, e.g., PERFECT \citep{karimi-mahabadi-etal-2022-prompt}. For fair comparison, we wrap the input with both soft and hard prompts and finetune both the PLM and the soft prompts on target-domain data. The predictions are mapped via verbalizers. 

\vspace{0.05in}
\noindent\textbf{5) Pre-trained Prompt Tuning (PPT).}
We follow \citet{gu-etal-2022-ppt}, who propose to transfer to sentence-pair classification tasks by pretraining on the next sentence prediction task with 10GB text from OpenWebText \citep{Gokaslan2019OpenWeb}. We download the pretrained checkpoint and finetune the soft prompt on the target domain directly. 

\vspace{0.05in}
\noindent\textbf{6) Soft Prompt Transfer (SPOT).}
\citet{vu-etal-2022-spot} propose to pretrain soft prompts on source-domain datasets and finetune the learned soft prompts on the target-domain datasets. We apply this approach on different source-target pairs in few-shot setting. 


\vspace{0.05in}
\noindent\textbf{7) Prompt Tuning with FreeLB.}
FreeLB \citep{Zhu2020FreeLB} is an adversarial training approach, which generates the adversarial perturbation from the supervised classification loss,
\begin{equation}
\bm \delta^* = \argmax_{\bm\delta,\|\bm\delta\|\leq\epsilon} \ell_{\text{xe}}(\bm x_s + \bm \delta, y_s, \bm p).
\end{equation}
After that, we find the optimal $\bm p$ by minimizing $\ell_{\text{xe}}(\bm x_s, y_s, \bm p) + \ell_{\text{xe}}(\bm x_s + \bm \delta, y_s, \bm p)$. The adversarial training may be understood as another type of smoothness constraint, as the network attempts to maintain the same prediction despite the strongest possible perturbation. 



\vspace{0.05in}
\noindent\textbf{8) Prompt Tuning with VAT.}
We apply the original VAT \citep{miyato2018virtual} to generate the perturbations that maximally alter model predictions on the source domain,
\begin{align}
\bm \delta^* & = \argmax_{\bm\delta,\|\bm\delta\|\leq\epsilon} \ell_{\text{KL}}(\bm \delta, \bm p, \bm x_s),
\label{eq:vat-baseline}
\end{align}
and optimize $\bm p$ as in Equation \ref{eq:tvat}. This can be seen as an ablation of \shortname{}, as Equation \ref{eq:vat-baseline} omits the $\ell_{\text{adv}}$ term from Equation \ref{eq:optimal-perturbation}. 

\vspace{0.05in}
\noindent\textbf{9) Prompt Tuning with DANN.} We implement Domain-adversarial Neural Network (DANN) \cite{dann-2016}, a popular UDA method for prompt tuning. DANN introduces a domain discrimination loss $\mathcal{L}_{\text{DD}}$,
\begin{align}
\mathcal{L}_{\text{DD}} =  \expect_{\bm{x}_s, \bm x_t} \big[ 
& - \mathrm{log} \ P(z=1|\bm x_s, \bm p) \\
& - \mathrm{log} \ P(z=0|\bm x_t, \bm p) \big], \nonumber
\end{align}
where $z$ is the output of a domain discrimination network. The soft prompt $\bm p$ optimizes for the source-domain cross-entropy loss and the negative domain discrimination loss.
\begin{equation}
\bm p^* = \argmin_{\bm p}\expect_{(\bm x_s, y_s)\in\mathcal{D}_s} \big[\ell_{\text{xe}}(\bm x_s, y_s, \bm p) \big] - \mathcal{L}_{\text{DD}}  
\end{equation}

For fair comparison, we use the same architecture for the domain discriminator as  \shortname{}. Note that in DANN, the gradients from domain discrimination loss are backpropagated to the soft prompts, while in \shortname{} such gradients are backpropagated to the perturbations.

\begin{table*}[ht]
\centering
\resizebox{\textwidth}{!}{%
\begin{tabular}{cccc|c|c|c|c|c}
\hline \Tstrut
\multirow{2}{*}{Method} &  \multirow{2}{*}{Params} & \multirow{2}{*}{PLM} & \multirow{2}{*}{Source} & \multicolumn{2}{c|}{QQP} & \multicolumn{2}{c|}{MRPC} & MNLI \\
& &  &  & \multicolumn{1}{c}{Acc.} & F1 & \multicolumn{1}{c}{Acc.} & \multicolumn{1}{c|}{F1} & Acc. \Bstrut\\
 \hline
 \Tstrut
 
Frozen & 0 & \multirow{5}{*}{T5-Large} & \xmark & 45.5 & 54.9 & 33.8 & \multicolumn{1}{c|}{11.8} & 41.7 \\

PT & 102K &  & \xmark & 48.4 $\pm$ 4.9 & 52.5 $\pm$ 5.5 & 53.1 $\pm$ 11.4 & 55.9 $\pm$ 23.4 & 33.4 $\pm$ 1.6 \\

FT & 770M &  & \xmark & 55.1 $\pm$ 6.7 & 52.0 $\pm$ 6.0 & \underline{59.5} $\pm$ 7.8 & \underline{67.9} $\pm$ 12.6 & 35.6 $\pm$ 2.4 \\

PFT & 770M &  & \xmark & \underline{55.1} $\pm$ 5.1 & \underline{57.8} $\pm$ 3.1 & 58.9 $\pm$ 11.0 & 65.3 $\pm$ 11.8 & 35.6 $\pm$ 3.6\\

PPT & 410K & T5-XXL & \cmark & 52.1 $\pm$ 11.1 & 56.2 $\pm$ 21.1 & 52.1 $\pm$ 11.1 & 56.2 $\pm$ 21.1 & 34.4 $\pm$ 1.4  \Bstrut\\

\hline\Tstrut
&&& & \multicolumn{2}{c|}{MRPC $\to$ QQP} & \multicolumn{2}{c|}{QQP $\to$ MRPC}  & SNLI $\to$ MNLI\\
& &  &  & \multicolumn{1}{c}{Acc.} & F1 & \multicolumn{1}{c}{Acc.} & \multicolumn{1}{c|}{F1} & Acc. \Bstrut\\

\hline\Tstrut
SPOT & 102K & \multirow{4}{*}{T5-Large} & \cmark & 64.5 $\pm$ 2.7 & 64.5 $\pm$ 0.8 & 68.7 $\pm$ 2.5 & 77.1 $\pm$ 2.9 & 74.3 $\pm$ 0.9\\

FreeLB & 102K &  & \cmark & 65.0 $\pm$ 2.4 & 64.5 $\pm$ 1.5 & 68.5 $\pm$ 2.2 & 77.6 $\pm$ 2.2 & 75.0 $\pm$ 1.0 \\

VAT & 102K &  & \cmark & 66.2 $\pm$ 2.0 & 64.9 $\pm$ 0.7 & 69.6 $\pm$ 1.9 & 79.0 $\pm$ 2.1 & 74.9 $\pm$ 1.1 \\

DANN & 102K &  & \cmark & 63.4 $\pm$ 2.5 & 62.5 $\pm$ 2.7 & 68.0 $\pm$ 3.5 & 76.2 $\pm$ 5.1 & 73.1 $\pm$ 1.4 \\

\shortname & 102K &  & \cmark & \textbf{69.1}* $\pm$ 1.7 & \textbf{65.8}* $\pm$ 1.9 & \textbf{71.2}* $\pm$ 1.7 & \textbf{79.9}* $\pm$ 1.7 & \textbf{78.4}* $\pm$ 0.6 \Bstrut\\

\hline\Tstrut

\multirow{2}{*}{Method} & \multirow{2}{*}{Params} & \multirow{2}{*}{PLM} & \multirow{2}{*}{Source} &  SNLI & \multicolumn{2}{c|}{SICK} & \multicolumn{2}{|c}{ CB} \\
& &  &  & Acc. & \multicolumn{2}{c|}{Acc.} & \multicolumn{2}{c}{Acc.}  \Bstrut\\
\hline\Tstrut

Frozen & 0 & \multirow{5}{*}{T5-Large} & \xmark  & 35.9 & \multicolumn{2}{c|}{37.1} & \multicolumn{2}{c}{55.4}\\

PT & 102K &  & \xmark & 34.6 $\pm$ 2.4 & \multicolumn{2}{c|}{61.5 $\pm$ 7.8} & \multicolumn{2}{|c}{38.3 $\pm$ 13.6}\\

FT & 770M &  & \xmark  & \underline{41.6} $\pm$ 3.8 & \multicolumn{2}{c|}{67.6 $\pm$ 6.3}  & \multicolumn{2}{c}{51.2 $\pm$ 7.8}\\

PFT & 770M &  & \xmark  & 38.6 $\pm$ 5.1 & \multicolumn{2}{c|}{\underline{71.3} $\pm$ 6.4} & \multicolumn{2}{|c}{\underline{57.3} $\pm$ 9.2} \\

PPT & 410K & T5-XXL & \cmark & 34.7 $\pm$ 2.8 & \multicolumn{2}{c|}{54.6 $\pm$ 14.0} & \multicolumn{2}{|c}{43.0 $\pm$ 14.6} \Bstrut\\
\hline\Tstrut

&& &  & MNLI $\to$ SNLI & SNLI $\to$ SICK & MNLI $\to$ SICK & SNLI $\to$ CB & MNLI $\to$ CB \Bstrut\\
& &  &  & Acc. & Acc. & Acc.  & Acc.  & Acc.   \Bstrut\\
\hline\Tstrut

SPOT & 102K & \multirow{4}{*}{T5-Large} & \cmark  & 78.8 $\pm$ 1.1 & 69.9 $\pm$ 5.3 & 72.9 $\pm$ 5.9 & 61.7 $\pm$ 5.0  & 65.3 $\pm$ 3.4\\

FreeLB & 102K &  & \cmark & 81.5 $\pm$ 0.7 & 69.5 $\pm$ 6.8 & 73.1 $\pm$ 4.8 & 61.6 $\pm$ 4.2  & 66.1 $\pm$ 3.3\\

VAT & 102K &  & \cmark  & 80.9 $\pm$ 0.9 & 68.6 $\pm$ 6.4 & 72.7 $\pm$ 6.3 & 59.0 $\pm$ 5.5 & 68.7 $\pm$ 4.8\\

DANN & 102K &  & \cmark & 71.1 $\pm$ 3.2 & 69.0 $\pm$ 6.7 & 73.4 $\pm$ 3.7 & 55.7 $\pm$ 5.5 & 66.9 $\pm$ 4.6 \\

\shortname & 102K &  & \cmark & \textbf{82.1}* $\pm$ 0.8 & \textbf{73.3} $\pm$ 6.8 & \textbf{74.8} $\pm$ 4.4 & \textbf{64.8}* $\pm$ 1.1  & \textbf{71.2}* $\pm$ 3.1 \Bstrut\\

\hline
\end{tabular}
}
\caption{Few-shot test performance. Results in bold are the best and results underlined are the best in the single-domain group. Results marked with * are significantly better than all the others under the student t-test ($p<0.05$).}
\label{tab:few-shot-results}
\end{table*}
 
\subsection{Experiment Settings}

\noindent\textbf{Pretraining.} For all methods that utilize source domain data, we train the soft prompts using the whole source-domain training set and perform model selection using the source-domain validation set. When domain adaptation is applied, we additionally use the entire target-domain training set for training with all labels removed. 
To mitigate variance, we train each method using 3 different random seeds, yielding three different models. For zero-shot evaluation, we report the mean score and standard deviation of the three models. 

\noindent\textbf{Few-shot Evaluation.} Following \citet{gao-etal-2021-making}, we sample the few-shot training set and validation set from the original target training set. Each set contains 8 data points per class. We evaluate the trained model on the original target validation set.
To mitigate high variance of few-shot learning, we repeat the sampling 16 times, and report the average of 48 runs (16 samples $\times$ 3 models). More details can be found in Appendix \ref{sec:appendix}.


\noindent\textbf{Model Settings.} For all the experiments, unless specified, we use the LM-adapted version of T5-large as the PLM. Results in \citet{lester-etal-2021-power} (Figure 3) shows that T5 further trained for LM Adaptation works the best for prompt tuning, which is also adopted by  \citet{gu-etal-2022-ppt} and \citet{vu-etal-2022-spot}. For the domain discriminator, we use a linear classification layer with parameters $\theta_d=[\bm w, \bm b], \bm w\in\mathbb{R}^{1024\times 2}, \bm b \in\mathbb{R}^{2}$, where $1024$ is the dimension of the output hidden states from the decoder of T5-large model.

\noindent\textbf{Soft and Hard Prompts.} Following \citet{lester-etal-2021-power,gu-etal-2022-ppt}, for all methods other than PPT, we set the soft prompt length to 100, initialized to the first 100 alphabetic token embeddings of T5. We combine soft prompts with hard prompts with details in the Appendix \ref{sec:appendix}.

\noindent\textbf{Evaluation Metrics.}
Following \citep{lester-etal-2021-power}, we use accuracy and F1 score to evaluate the performance on the MRPC and QQP datasets. Following \citep{gu-etal-2022-ppt}, we use accuracy for NLI. For zero-shot model selection, we use the source-domain validation set. For few-shot model selection, we use the target-domain validation set. 


\begin{table*}[ht]
\centering
\resizebox{\textwidth}{!}{%
\begin{tabular}{c||c||c|c||c||c|c|c}
\hline \Tstrut
\multirow{2}{*}{Method} &  MRPC & \multicolumn{2}{c||}{MRPC $\to$ QQP} &  QQP & \multicolumn{2}{c|}{QQP $\to$ MRPC} & MNLI $\to$ CB \\
& Acc. & \multicolumn{1}{c}{Acc.} & F1 & Acc.  & \multicolumn{1}{c}{Acc.} & \multicolumn{1}{c|}{F1} & \multicolumn{1}{c}{Acc.} \Bstrut\\
\hline\Tstrut

SPOT &   82.5 $\pm$ 1.5 &    60.9 $\pm$ 4.6 &   63.6 $\pm$ 2.0 &   80.9 $\pm$ 2.2 &  65.7 $\pm$ 3.4 &  73.2 $\pm$ 5.7 & 63.2 $\pm$ 5.7\\

FreeLB &    85.5 $\pm$ 0.3 &  63.1 $\pm$ 3.7 &  63.9 $\pm$ 1.0 &  82.2 $\pm$ 2.7 &  69.4 $\pm$ 1.1 &  78.7 $\pm$ 1.3 & 67.8 $\pm$ 3.9 \\

VAT &   84.7 $\pm$ 0.8 &  64.8 $\pm$ 4.6 &  64.1 $\pm$ 1.7 &  81.9 $\pm$ 0.7 &    68.9 $\pm$ 1.5 &  78.5 $\pm$ 1.5 & 67.8 $\pm$ 5.8\\

DANN & 81.5 $\pm$ 2.1 & 63.9 $\pm$ 1.8 & 57.6 $\pm$ 3.3 & 81.4 $\pm$ 0.7 & 63.6 $\pm$ 4.8 & 71.5 $\pm$ 9.7 & 59.8 $\pm$ 4.4 \\

\shortname & \textbf{85.7} $\pm$ 0.7 & \textbf{68.9} $\pm$ 0.8 & \textbf{66.3} $\pm$ 0.6 & \textbf{82.7} $\pm$ 1.3 & \textbf{ 71.2} $\pm$ 0.4 & \textbf{80.0} $\pm$ 0.6 & \multicolumn{1}{c}{\textbf{68.3} $\pm$ 2.6} \Bstrut\\

\hline \Tstrut

\multirow{2}{*}{Method} & MNLI & MNLI $\to$ SNLI & MNLI $\to$ SICK & SNLI & SNLI $\to$ MNLI & SNLI $\to$ SICK & SNLI $\to$ CB \\
& Acc. & Acc. & Acc. & Acc. & Acc. & Acc. & \multicolumn{1}{c}{Acc.} \Bstrut\\
\hline\Tstrut

SPOT & 83.4 $\pm$ 0.8 & 79.2 $\pm$ 1.0 & 51.8 $\pm$ 0.7 &  88.9 $\pm$ 0.1 &    75.6 $\pm$ 0.4 & 52.7 $\pm$ 1.9  &  47.6 $\pm$ 3.7 \\

FreeLB & \textbf{84.8} $\pm$ 0.8 & 81.8 $\pm$ 0.7 & 52.2 $\pm$ 0.2 &   \textbf{89.9} $\pm$ 0.1 &  77.5 $\pm$ 0.5  &  52.9 $\pm$ 1.9 & 47.5 $\pm$ 4.7 \\

VAT & 83.7 $\pm$ 0.3 & 81.0 $\pm$ 0.2 & 51.4 $\pm$ 1.4 & 88.7 $\pm$ 0.1  &  77.1 $\pm$ 1.3  & 51.8 $\pm$ 2.1 & 45.8 $\pm$ 0.8 \\

DANN & 80.4 $\pm$ 2.7 & 72.4 $\pm$ 5.9 & \textbf{61.9} $\pm$ 2.7 & 85.3 $\pm$ 3.2 & 70.3 $\pm$ 3.6 & 51.5 $\pm$ 1.2 & 42.3 $\pm$ 2.2 \\

\shortname & 84.6 $\pm$ 0.3 & \textbf{82.1} $\pm$ 0.8 & 55.2 $\pm$ 1.0 &    89.2 $\pm$ 0.1  &   \textbf{79.1} $\pm$ 0.1  & \textbf{ 53.8} $\pm$ 0.5 & \textbf{49.4} $\pm$ 4.2 \Bstrut\\

\hline
\end{tabular}
}
\caption{Source-domain and zero-shot target-domain test performance.}
\label{tab:zero-shot-results}
\end{table*}

\subsection{Few-shot Performance} 
\label{exp:few-shot-with-init}
We adopt few-shot classification to evaluate the representations learned by different models and pretraining methods. 
We show the few-shot performance in Table \ref{tab:few-shot-results} and make the following observations. First, \shortname{} significantly outperforms all baseline models across all the few-shot test cases, including the state-of-the-art SPOT baseline. We perform statistical significance tests that compare \shortname{} to all baselines in a pair-wise manner. 
In all but the SICK experiments, the differences between \shortname{} and all baselines are statistically significant. We attribute the performance to the high-quality representation of \shortname{}, resulting from domain adaptation. 

Second, DANN performs much worse than perturbation-based methods. As discussed earlier, we suspect the poor performance of DANN is partially due to the limited capacity of prompts (102K parameters in our case). In OPTIMA, the perturbation optimizes for domain invariance (Eq. \ref{eq:optimal-perturbation}), whereas the prompt optimizes for only task-specific losses (Eq. \ref{eq:tvat}), which simplifies optimization for soft prompts.


Third, \shortname{} outperforms the VAT baseline, especially in the NLI tasks, where the performance difference ranges from 1.2\% in MNLI$\rightarrow$SNLI to 5.8\% in SNLI$\rightarrow$CB. The VAT baseline is an ablation of \shortname{} and omits the targeted regularization term when finding the perturbation. This comparison demonstrates the effectiveness of the proposed targeted smoothness constraint. 

Finally, our experiments are consistent with earlier results of \citet{gu-etal-2022-ppt}, which show that prompt tuning (PT) suffers from high variance in the results. In the single-domain experiments, finetuning the entire T5-Large (FT) exhibits comparable, if not lower, variances than PT, even though FT updates about 7500$\times$ more parameters. This underscores the importance of using pretrained prompts from a source domain. Indeed, all transfer learning methods utilizing a source domain similar to the target (SPOT, FreeLB, VAT, and \shortname{}) yield sizable performance gains than single-domain methods. Notably, FreeLB, VAT and \shortname{} are obviously better than SPOT across the benchmarks, which underscores the importance of alleviating overfitting to source-domain datasets. 

\begin{figure}[!t]
    \centering
\includegraphics[width=\columnwidth]{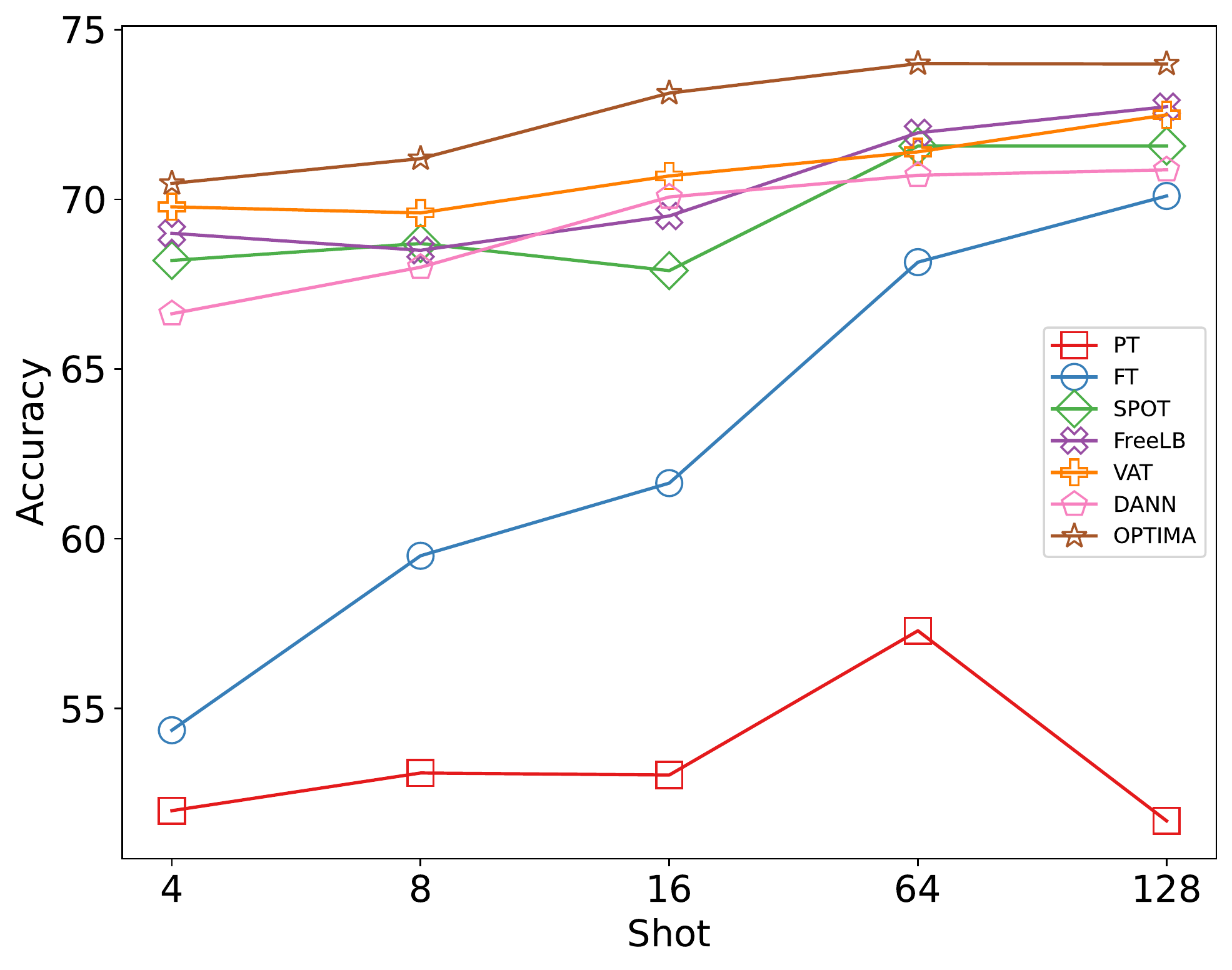}
    \caption{Average test performance on MRPC across 16 runs. PT and FT are trained on MRPC directly. The rest use the soft prompt pretrained under the QQP$\rightarrow$MRPC setting as initialization. \shortname{} exhibits the best performance across different few-shot settings. }
    \label{fig:more-shots}
\end{figure}

\noindent\textbf{Sample Efficiency.} 
We perform an additional experiment where we increase the number of available samples per class from the target domain, and show the results in Figure \ref{fig:more-shots}. We observe that 4-shot \shortname{} achieves comparable performance as full-model finetuning on 128-shot dataset. Similarly, 8-shot \shortname{} achieves an accuracy  comparable to 64-shot SPOT. These results clearly demonstrate the superior sample efficiency of \shortname. 

\begin{figure}[!t]
    \centering
\includegraphics[width=\columnwidth]{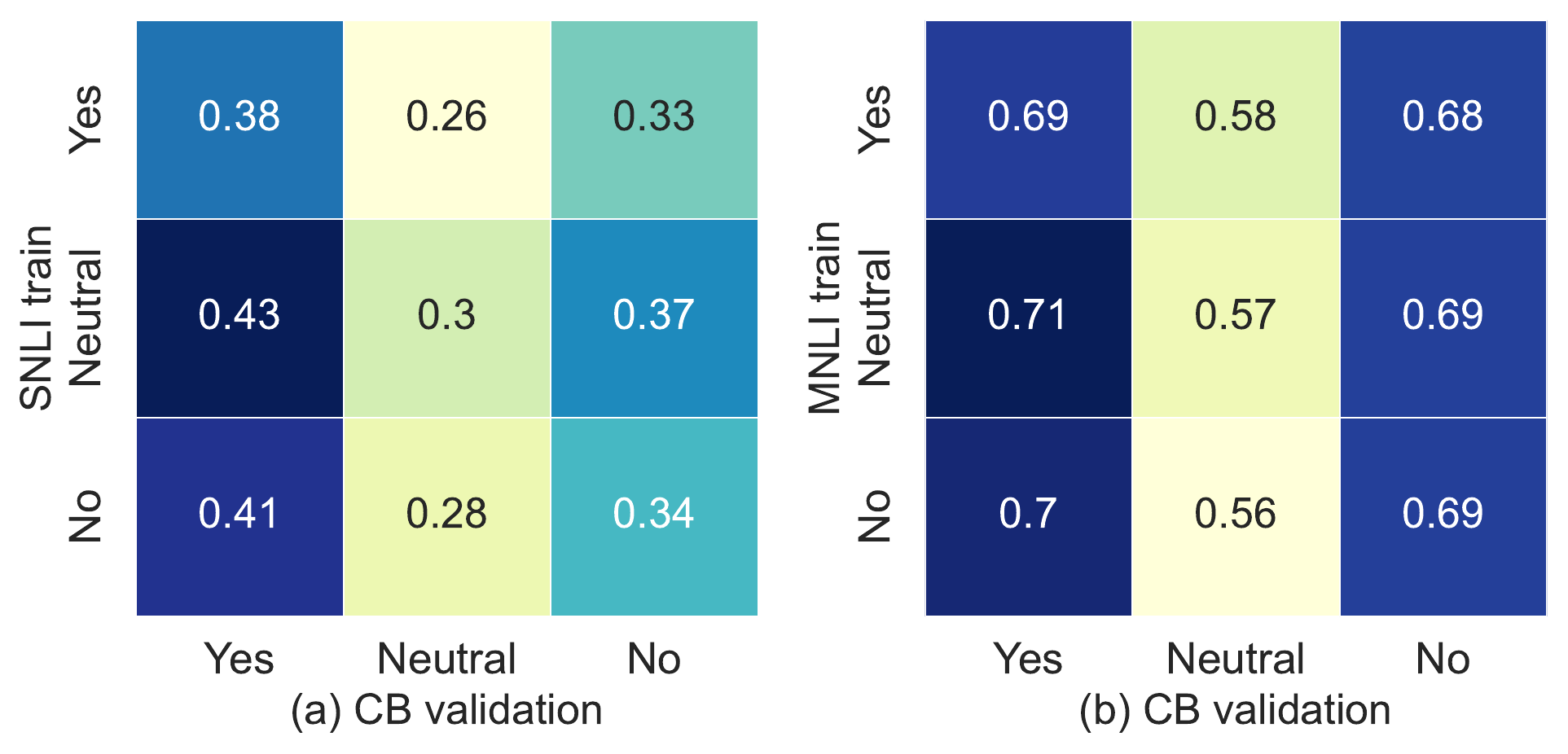}
    \caption{TF-IDF similarity for SNLI, MNLI, and CB, where we treat all text in one class as a document. }
    \label{fig:document_similarity}
\end{figure}

\subsection{Zero-shot Performance}
Zero-shot performance on the target domain is also an effective way to evaluate the learned representations. We show the zero-shot performance in Table \ref{tab:zero-shot-results} and make the following observations. 

First, \shortname{} still takes the highest spot in performance in all target domains, outperforming the second best baseline by up to 4.1\%. In the source domain, \shortname{} is comparable with the baselines. Second, the ablation baseline, VAT, is consistently surpassed by \shortname{}, which again confirms the utility of our proposal. Third, the state-of-the-art method, SPOT, in the majority of cases produces results with higher variance than the three perturbation-based methods. This suggests that adversarial perturbation is effective against overfitting. Lastly, except in the MNLI $\rightarrow$ SICK task, DANN performs rather poorly across the benchmarks, indicating that DANN is not suitable for prompt tuning.




\subsection{Class Similarity and Transfer Learning}
\label{exp:class-level-study}

We investigate the relationship between domain similarity and transfer learning performance. Due to space constraints, we present the results on CB as the target domain and leave more content to the Appendix. CB is a difficult target. On SNLI, all models in Table \ref{tab:zero-shot-results} achieve in-domain test accuracy greater than $88\%$, but zero-shot SNLI-to-CB transfer obtains accuracy of around $47\%$. This is disappointing given that even Frozen PLM achieves $55.4\%$ on CB. 

\begin{figure}[!t]
    \centering
\includegraphics[width=\columnwidth]{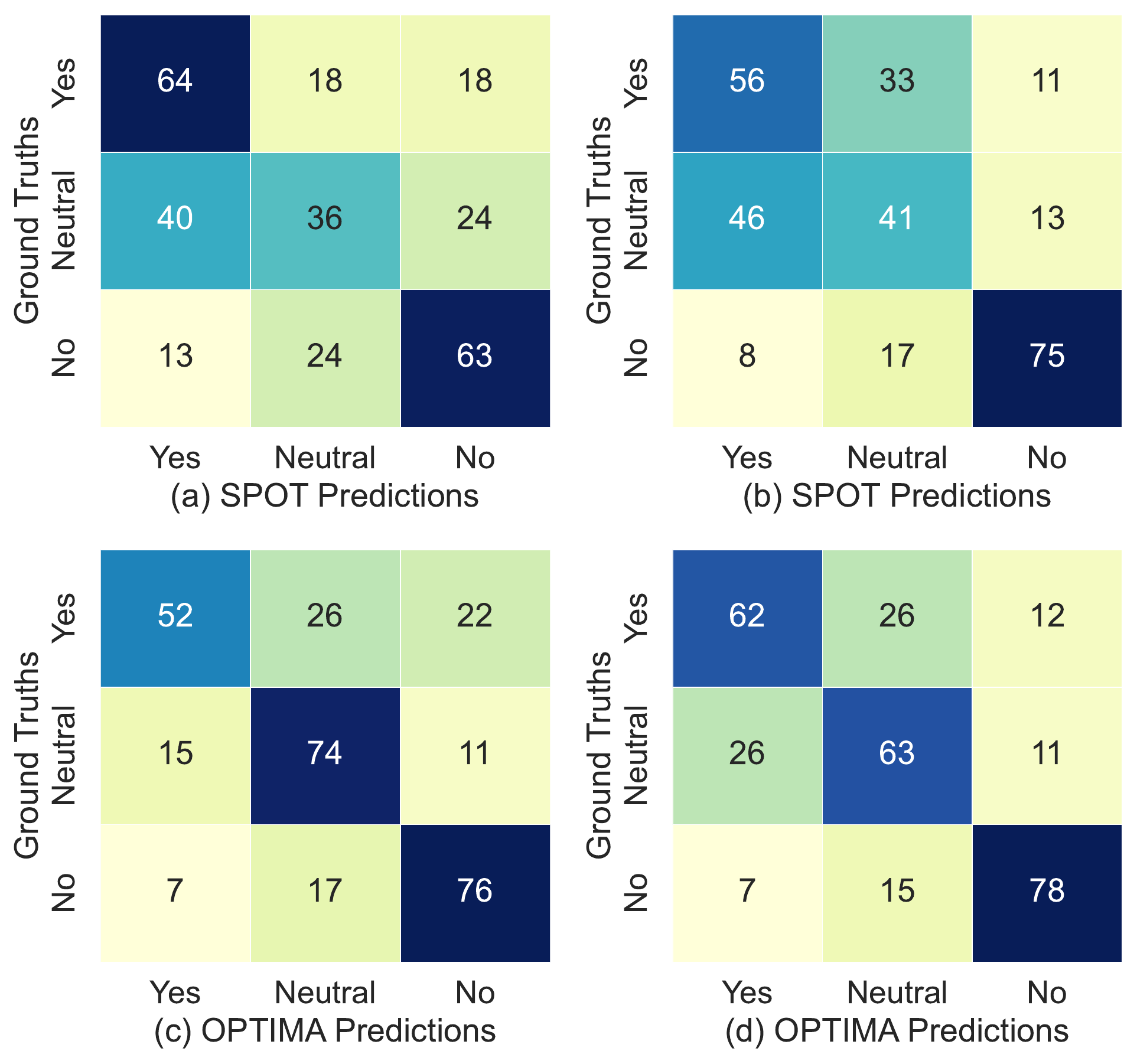}
    \caption{Confusion matrices for 8-shot transfer learning to CB. Each result is the average test accuracy across 16 runs. (a) and (c) refer to SNLI$\rightarrow$CB while (b) and (d) refer to MNLI$\rightarrow$CB setting respectively}
    \label{fig:confusion-matrix}
\end{figure}

\begin{figure}[ht]
    \centering
\includegraphics[width=\columnwidth]{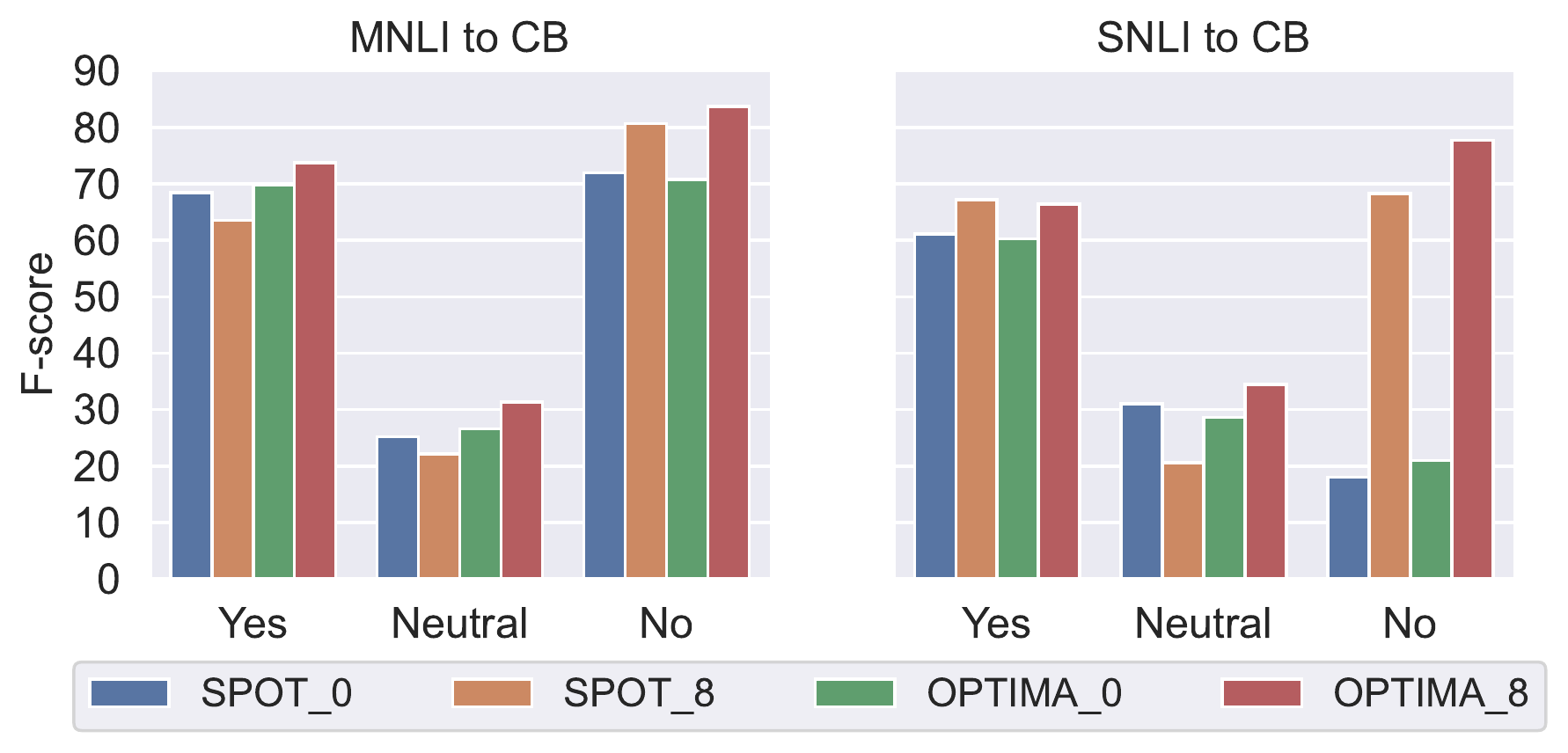}
    \caption{Class-level F1-score on the CB datasets. $\textrm{SPOT}\_0$ and $\textrm{\shortname}\_0$ denote zero-shot performance. $\textrm{SPOT}\_8$ and $\textrm{\shortname}\_8$ denote 8-shot performance.}
    \label{fig:nli_classes}
\end{figure}

To investigate the underlying cause, we plot the TF-IDF textual similarities between different domains in Figure \ref{fig:document_similarity}. We compare SPOT, which performs direct transfer without any smoothness regularization, and \shortname{} in the form of confusion matrices in Figure \ref{fig:confusion-matrix} and F1 scores in Figure \ref{fig:nli_classes}.  

Figure \ref{fig:document_similarity}(a) shows irregular similarities between classes of SNLI and CB, which explains the difficulty in transfer learning. For example, the SNLI Neutral class is more similar to the CB Yes class than the CB Neural class. The CB Neutral class has low similarity to all SNLI classes. This leads to significant confusion for the few-shot SPOT classifier in the SNLI-to-CB transfer and especially low accuracy for the CB Neutral class (Figure \ref{fig:nli_classes}). The situation is similar for the MNLI-to-CB transfer. Interestingly, the regularization of \shortname{} is able to alleviate the domain shift and obtain accuracy improvements for the CB No and Neutral classes.



\section{Related Work}
\noindent\textbf{Few-shot Learning with PLMs.}
Traditional approach for few-shot learning is \textit{fine-tuning}, where a PLM and a task-specific head are tuned together for the tasks at hand \citep{zhang-etal-2021-effectiveness-pre,chen-etal-2020-shot,das-etal-2022-container}. However, fine-tuning causes high memory consumption as the scale of PLMs increases. To better exploit large frozen PLMs, prompt-based methods have demonstrated excellent few-shot performance on a range of datasets by wrapping test examples in cloze question format for GPT-3 to make predictions  \citep{brown2020language}. Prompts are also shown to boost fine-tuning in LM-BFF \citep{gao-etal-2021-making}, PET \citep{schick-schutze-2021-exploiting,schick-schutze-2021-just}, and PERFECT \citep{karimi-mahabadi-etal-2022-prompt}.

\noindent\textbf{Transfer Learning for Prompt Tuning.}
Soft prompt tuning methods \citep{lester-etal-2021-power,li-liang-2021-prefix,liu-etal-2022-p,hambardzumyan-etal-2021-warp} learn prompts from data and achieve comparable performance with full-model tuning when the PLMs are large enough. SPOT \citep{vu-etal-2022-spot} proposes to pretrain soft prompts on a set of source-domain datasets and then use the trained soft prompts to boost prompt tuning for target domains. PPT \citep{gu-etal-2022-ppt} introduces unsupervised tasks such as next sentence prediction as the pre-text task for prompt pretraining.  After that, the soft prompts are finetuned on the few-shot target-domain data. \citet{wang-etal-2021-transprompt} pretrain soft prompts across few-shot datasets. Different from these methods, we explore the use of unlabeled target domain data in few-shot prompt tuning. 

\noindent\textbf{Consistency Training for NLP.} Consistency training methods \citep{conf/iclr/LaineA17,NIPS2016_30ef30b6,wei-zou-2019-eda,ng-etal-2020-ssmba,xie2020unsupervised} force the model to make consistent predictions against small perturbations. For example, \citet{park2021consistency} produce discrete virtual adversarial noise to the token embeddings. \citet{yoon-etal-2021-ssmix} apply mixup to perturb the spans of the input texts for text classification. \citet{kim-etal-2021-learn} propose a consistency training framework to enhance the conversational dependency of question answering. Different from these single-domain settings, we consider a cross-domain setting and exploit domain adaptation for regularization. 

\noindent\textbf{Neural Domain Adaptation for NLP.} Neural domain adaptation \cite{ben2010theory} includes supervised domain adaptation (SDA) \cite{zhou-etal-2019-dual} and unsupervised domain adaptation (UDA) \cite{wang-etal-2019-adversarial,long-etal-2022-domain}, depending on whether the target-domain data are labeled or unlabeled. Domain adaptation has been used in various applications such as sentiment analysis \cite{glorot2011domain,dai2020adversarial,ghosal-etal-2020-kingdom}, machine translation \cite{chu2017empirical}, reading comprehension \cite{wang-etal-2019-adversarial}, and others \cite{shah-etal-2018-adversarial,naik-2020-event-dann}. For a complete survey of UDA in NLP, we refer readers to \citet{ramponi-plank-2020-neural}. In this paper, we do not induce domain-invariant soft prompts but encourage the learned adversarial perturbations to fill in the domain gap and focus on smoothing the decision boundary where source-domain and target-domain data are similar.

\section{Conclusions}
In this paper, we propose \shortname{} to enhance soft prompt transfer performance by regularizing the training on source domain under perturbations generated with domain adaptation. We extensively evaluate the proposed method. Compared to  competitive baselines, soft prompt trained with \shortname{} generalizes better to the source domain and significantly boosts zero-shot and few-shot learning in the target domain. We observe that pre-training soft prompts on a similar dataset confers more benefits than pre-training on a disimilar dataset. We expect the current work to contribute to the wide deployment of PLMs.

\section*{Acknowledgments}
This research is supported by the National Research Foundation (NRF), Singapore under its AI Singapore Programme (AISG2-RP-2020-019); the Joint NTU-WeBank Research Centre on Fintech (NWJ-2020-008); the Nanyang Assistant/Associate Professorship (NAP); the RIE 2020 Advanced Manufacturing and Engineering (AME) Programmatic Fund (A20G8b0102), Singapore; Future Communications Research \& Development Programme (FCP-NTU-RG-2021-014); and NRF Fellowship (NRF-NRFF13-2021-0006). Any opinions, findings, conclusions, or recommendations expressed in this material are those of the authors and do not reflect the views of 
the funding agencies.

\section*{Limitations}
We identify a few limitations of the current work. 
\begin{itemize}
\item The domain adaptation problem formulation requires unlabeled data from the target domain. Although unlabeled data are easy to obtain in most cases, doing so might be difficult for some data-scarce domains. 

\item The proposed regularization technique addresses the situation where the source and target domains have different data distributions. When the two distributions are exactly the same, the technique degenerates to simply adversarial training. When the two distributions are extremely dissimilar, the transfer is unlikely to yield performance improvements. A unified framework that automatically detects domain distances and applies the correct method may be desirable. 

\item  The power of perturbations has the most effect in the few-shot / zero-shot settings. When the target domain has abundant labeled data, the gap between soft prompt tuning and our method will likely diminish. 
\end{itemize}




\bibliography{anthology,custom}
\bibliographystyle{acl_natbib}

\newpage
\clearpage
\appendix

\section{Appendix}
\label{sec:appendix}

\subsection{Few-shot Evaluation Protocol}
In PET \citep{schick-schutze-2021-just}, authors evaluated its few-shot performance using a fixed training set. In LM-BFF \citep{gao-etal-2021-making}, authors conducted more studies on the configuration of few-shot settings and proposed to average 5 randomly sampled few-shot sets. We determine the sample size, 16, based on an statistical analysis\footnote{\url{https://www.itl.nist.gov/div898/handbook/prc/section2/prc222.htm}} on the sample size required for investigating an unknown population mean under student $t$-test. Here we adopt a significance level $\alpha=0.05$, the risk of rejecting a true hypothesis that the performance of one method is better than the other.

For all the cross-domain few-shot learning methods, the few-shot test performance of 3 differently pre-trained soft prompts are averaged for each given $\mathcal{D}_{train}$ and $\mathcal{D}_{dev}$ splits, and we obtain 16 averaged few-shot performance. Then we compute the mean and standard deviation for the 16 test results.

\noindent\textbf{Training Settings.} Following \citep{lester-etal-2021-power}, we use Adafactor \citep{shazeer2018adafactor} as the optimizer and set the learning rate to $0.3$ for all the pre-training tasks on the entire source domain dataset. We use the cosine learning rate scheduler for all methods. For the pre-training stage, we set the maximum number of training steps to $30,000$ and evaluate the models on the validation set every $1,000$ steps. We set the batch size to $8$ for MRPC and QQP, and $18$ for the NLI datasets. For the few-shot learning setting, we set the maximum number of training steps to $1,000$ and evaluate models on $|\mathcal{D}_{dev}|$ every $4$ steps. we set batch size to $4$ for MRPC and QQP, and $6$ for the NLI datasets. All the training are done on NVIDIA V-100 with 32 GB.

\begin{table}[ht]
\centering
\resizebox{\columnwidth}{!}{%
\begin{tabular}{c|c}
\hline 
& Hybrid Template \\
\hline
T1 & $\bm P$ $<S_1>$ and $<S_2>$ are equivalent? $\textrm{[MASK]}$\\
\hline
T2 & $\bm P$ hypothesis: $<S_1>$ premise: $<S_2>$ answer: $\textrm{[MASK]}$ \Bstrut\\
\hline
\end{tabular}
}
\caption{The hybrid templates where $\bm P$ represents learnable soft prompts. $<S_1>$ and $<S_2>$ are sentence pairs. $\textrm{[MASK]}$ represents the labels to be predicted. T1 is the template adopted by the paraphrase detection and question pair classification tasks. T2 is the template adopted by four natural language inference tasks.}
\label{tab:template}
\end{table}

\begin{table}[t]
\centering
\resizebox{\columnwidth}{!}{%
\begin{tabular}{@{}cccccc@{}}
\hline\Tstrut
Methods & 4-shot & 8-shot & 16-shot & 64-shot & 128-shot \Bstrut\\
\hline\Tstrut
PT & 51.9 $\pm$ 8 & 53.1 $\pm$ 11 & 53.0 $\pm$ 9 & 57.2 $\pm$ 9 & 51.7 $\pm$ 10 \\
FT & 54.4 $\pm$ 12 & 59.5 $\pm$ 8 &  61.6 $\pm$ 6 &  68.2 $\pm$ 4 & 70.1 $\pm$ 5 \Bstrut\\
\hline \Tstrut
SPOT & 68.2 $\pm$ 4 & 68.7 $\pm$ 3 & 67.9 $\pm$ 5 & 71.6 $\pm$ 4 & 71.6 $\pm$ 4 \\
FreeLB & 69 $\pm$ 4 & 68.5 $\pm$ 2 & 69.5 $\pm$ 2.5 & 71.9 $\pm$ 1.7 & 72.7 $\pm$ 1.8 \\
VAT & 69.8 $\pm$ 2.7 & 69.6 $\pm$ 1.9 & 70.7 $\pm$ 2.7 & 71.4 $\pm$ 3.1 & 72.5 $\pm$ 2.9  \\
DANN & 66.6 $\pm$ 6.2 & 63.6 $\pm$ 4.8  & 70.1 $\pm$ 3.8 & 70.7 $\pm$ 3.8 & 70.9 $\pm$ 2.2 \\
\shortname{} & 70.5 $\pm$ 3.4 & 71.2 $\pm$ 1.7 & 73.1 $\pm$ 2.1 & 74 $\pm$ 2.7 & 74 $\pm$ 2.1  \Bstrut\\
\hline
\end{tabular}
}
\caption{Average test performance on MRPC dataset, where transfer learning methods are transferred from the QQP dataset.}
\label{tab:sample-efficiency}
\end{table}

\begin{table}[t]
\resizebox{\columnwidth}{!}{%
\centering
\begin{tabular}{c|c|ccc|ccc}
\hline
\multicolumn{2}{c}{\multirow{2}{*}{}} & \multicolumn{3}{c}{SNLI} & \multicolumn{3}{|c}{MNLI} \\
\hline
\multicolumn{2}{c}{} & \multicolumn{1}{|c}{Yes} & Neutral & No & Yes & Neutral & No \\
\hline
\multirow{3}{*}{CB} & Yes & \textbf{83.7} & 7.61 & 8.7 & \textbf{70.43} & 20.87  & 8.7 \\

 \hline
 & Neutral & \textbf{65} & 20 & 15 & \textbf{56} & 44 & 0 \\
 \hline
 & No & \textbf{62.5} & 12.5 & 25 & 19.29 & 20  & \textbf{60.71}\\
 \hline
\end{tabular}%
}
\caption{Confusion matrix for \textbf{zero-shot} performance of \textbf{SPOT} on each class of $\textrm{CB}$. Results are in $\%$. The bold means the most predicted labels for each of the classes of $\textrm{CB}$ }
\label{tab:cb-spot-zero-cmtrix}
\end{table}

\begin{table}[t]
\resizebox{\columnwidth}{!}{%
\centering
\begin{tabular}{c|c|ccc|ccc}
\hline
\multicolumn{2}{c}{\multirow{2}{*}{}} & \multicolumn{3}{c}{SNLI} & \multicolumn{3}{|c}{MNLI} \\
\hline
\multicolumn{2}{c}{} & \multicolumn{1}{|c}{Yes} & Neutral & No & Yes & Neutral & No \\
\hline
\multirow{3}{*}{CB} & Yes & \textbf{76.52} & 18.26 & 6.52 & \textbf{71.3} &  20  & 8.7\\

 \hline
 & Neutral & \textbf{60} & 40 & 0 & \textbf{52} & 48 & 0 \\
 \hline
 & No & \textbf{52.86} & 29.29 & 17.86 & 17.86 & 20.71 & \textbf{61.43}  \\
 \hline
\end{tabular}%
}
\caption{Confusion matrix for \textbf{zero-shot} performance of \textbf{\shortname} on each class of $\textrm{CB}$.}
\label{tab:cb-ours-zero-cmtrix}
\end{table}

\begin{table}[t]
\resizebox{\columnwidth}{!}{%
\centering
\begin{tabular}{c|c|ccc|ccc}
\hline
\multicolumn{2}{c}{\multirow{2}{*}{}} & \multicolumn{3}{c}{SNLI} & \multicolumn{3}{|c}{MNLI} \\
\hline
\multicolumn{2}{c}{} & \multicolumn{1}{|c}{Yes} & Neutral & No & Yes & Neutral & No  \\
\hline
\multirow{3}{*}{CB} & Yes & \textbf{63.86} & 17.93 & 18.21 & \textbf{56.25} & 32.61  & 11.14 \\
 \hline
 & Neutral & \textbf{40} & 36.25 & 23.75 & \textbf{46.25} & 41.25 & 12.5 \\
 \hline
 & No & 12.95 & 23.88 & \textbf{63.17} & 8.04 & \textbf{16.52} & 75.45 \\
 \hline
\end{tabular}%
}
\caption{Confusion matrix for \textbf{few-shot} performance of \textbf{SPOT} on each class of $\textrm{CB}$. }
\label{tab:cb-spot-few-cmtrix}
\end{table}

\begin{table}[t]
\resizebox{\columnwidth}{!}{%
\centering
\begin{tabular}{c|c|ccc|ccc}
\hline
\multicolumn{2}{c}{\multirow{2}{*}{}} & \multicolumn{3}{c}{SNLI} & \multicolumn{3}{|c}{MNLI} \\
\hline
\multicolumn{2}{c}{} & \multicolumn{1}{|c}{Yes} & Neutral & No & Yes & Neutral & No \\
\hline
\multirow{3}{*}{CB} & Yes & \textbf{52.45} & 25.54 & 22.01 & \textbf{61.68} & 26.09  & 12.23 \\
 \hline
 & Neutral & 15 & \textbf{73.75}  & 11.25 & 26.25 & \textbf{62.5}  & 11.25 \\
 \hline
 & No & 7.37 & 16.96 & \textbf{75.67}  & 6.69 & 14.96 & \textbf{78.35} \\
 \hline
\end{tabular}%
}
\caption{Confusion matrix for \textbf{few-shot} performance of \textbf{\shortname} on each class of $\textrm{CB}$. }
\label{tab:cb-ours-few-cmtrix}
\end{table}

\begin{figure}[t]
    \centering
\includegraphics[width=\columnwidth]{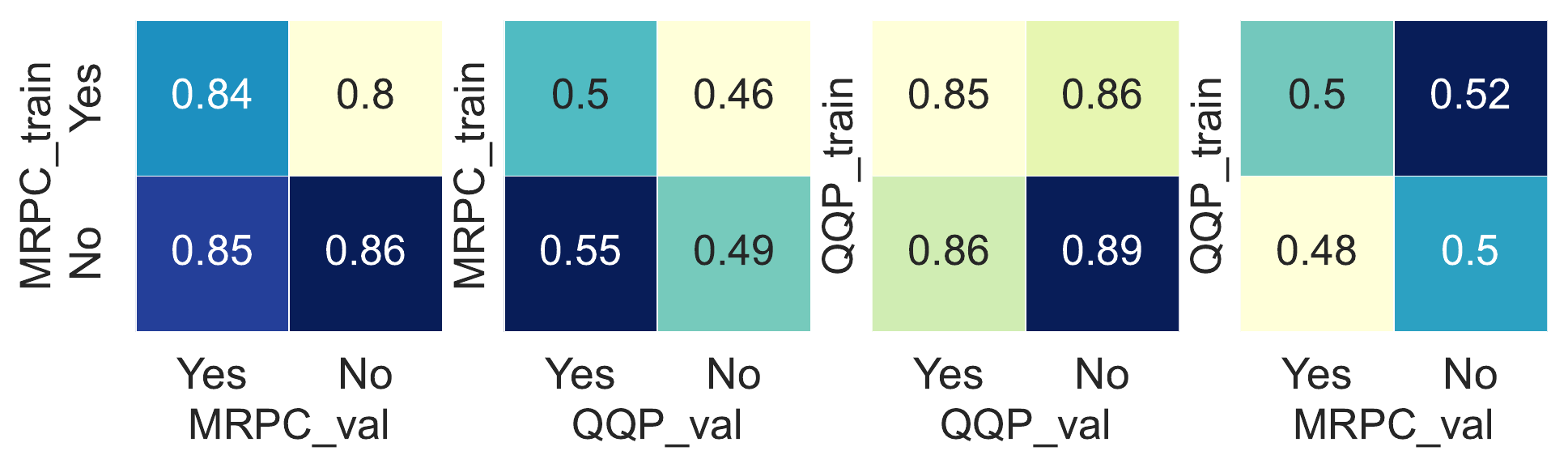}
    \caption{Document similarity for $\textrm{MRPC}$ and $\textrm{QQP}$ datasets between their classes. }
    \label{fig:document_similarity_qqp}
\end{figure}

\begin{figure}[t]
    \centering
\includegraphics[width=\columnwidth]{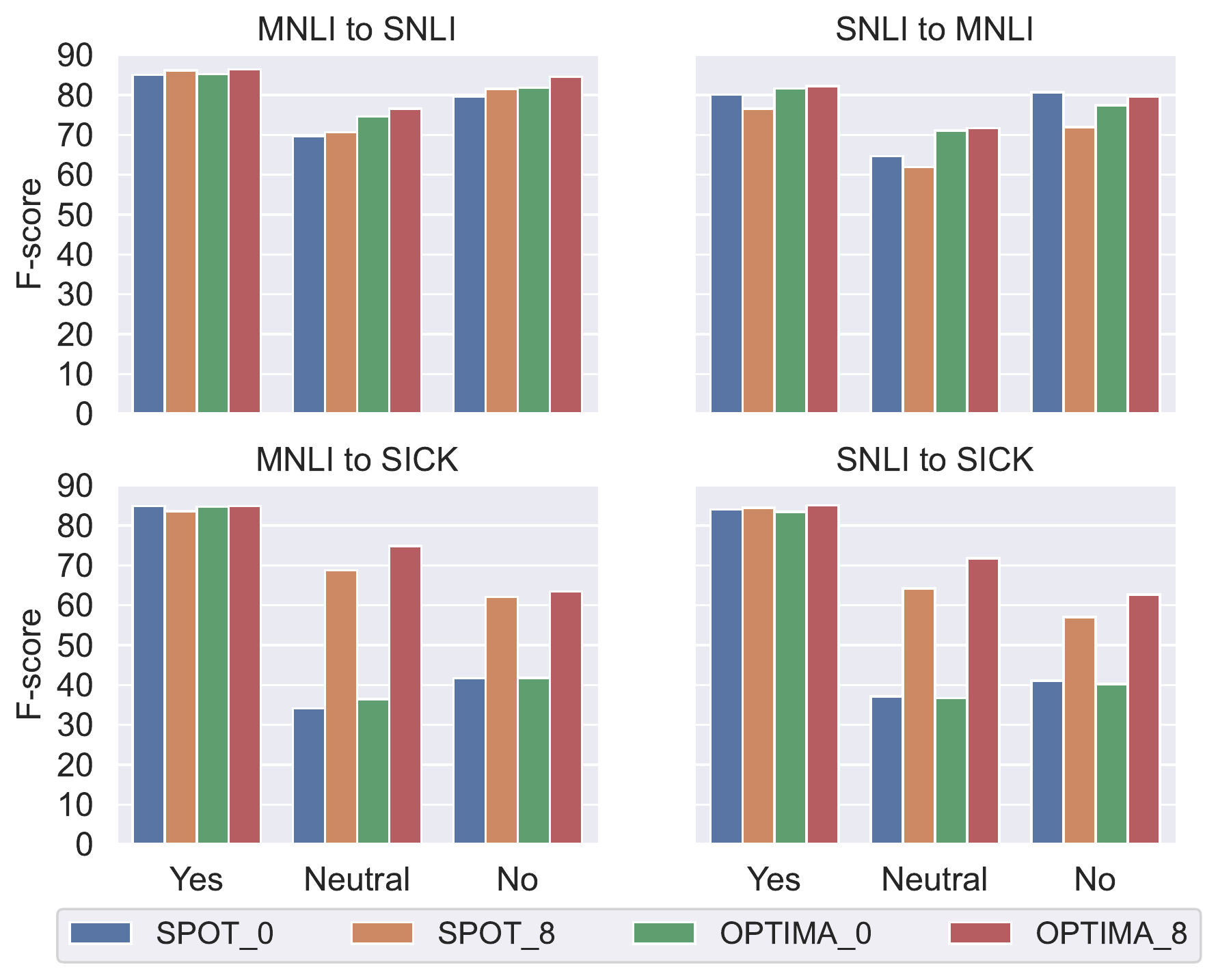}
    \caption{F-score on three classes for NLI datasets. $\textrm{SPOT}\_0$ and $\textrm{\shortname}\_0$ are compared for their zero-shot performance. $\textrm{SPOT}\_8$ and $\textrm{\shortname}\_8$ are compared for their 8-shot performance.}
    \label{fig:nli_classes_more}
\end{figure}

\begin{figure}[t]
    \centering
\includegraphics[width=\columnwidth]{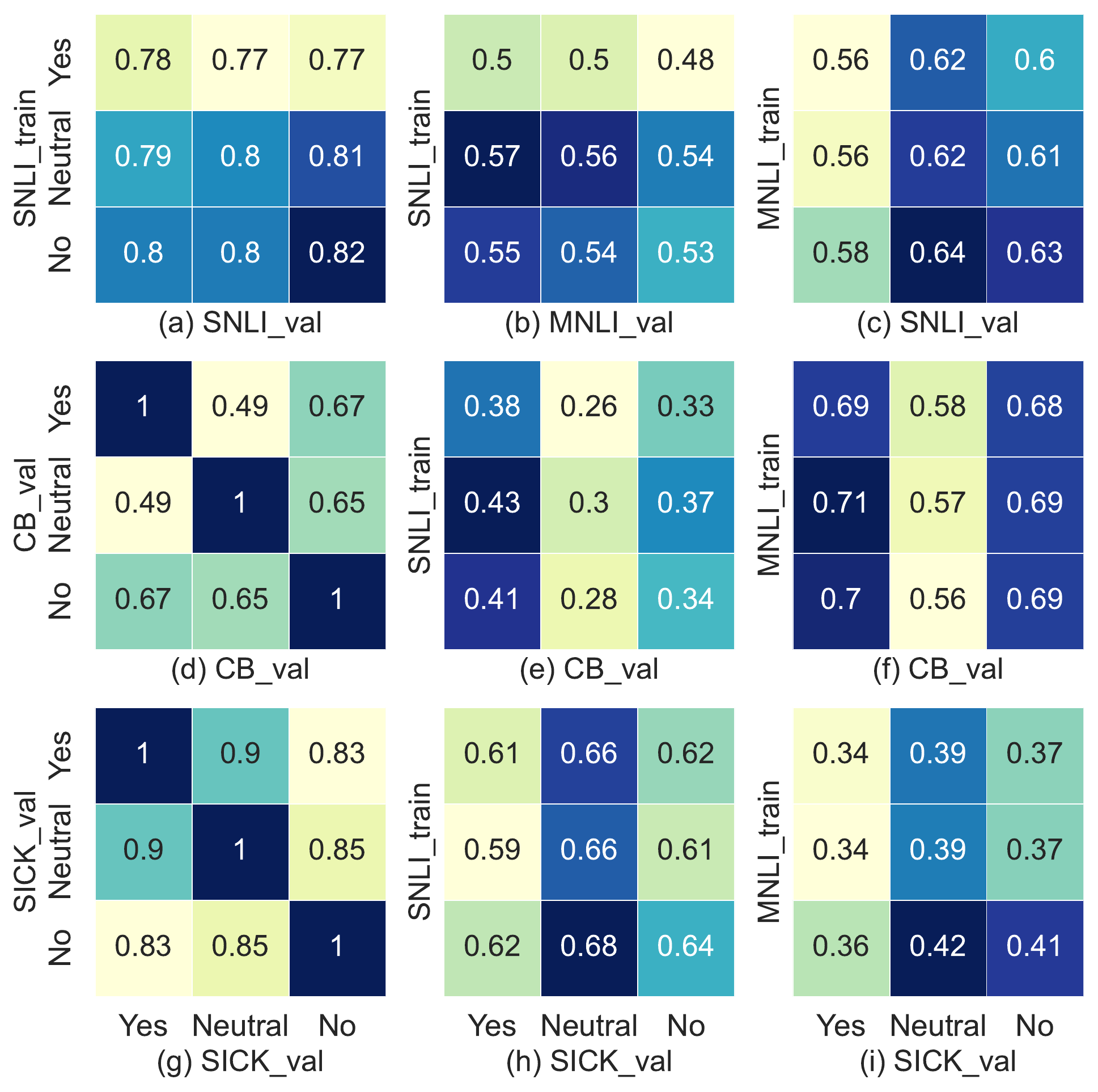}
    \caption{Document similarity using TF-IDF for each pair of NLI datasets. }
    \label{fig:document_similarity_more}
\end{figure}

\end{document}